\documentclass{article}

\usepackage{microtype}
\usepackage{graphicx}
\usepackage{subfigure}
\usepackage{booktabs} % for professional tables
\usepackage{amsmath}
\usepackage{amssymb}
\usepackage{mathtools}

\usepackage[accepted]{icml2019}

\newcommand{\pp}[1]{\left( #1 \right)}

\definecolor{darkgreen}{rgb}{0,0.6,0}
\definecolor{darkred}{rgb}{0.7,0.0,0}
\definecolor{darkblue}{rgb}{0,0.0,0.6}

\icmltitlerunning{Using learned optimizers to make models robust to input noise}

\begin{document}

\twocolumn[

\icmltitle{Using learned optimizers to make models robust to input noise}

\icmlsetsymbol{equal}{*}

\begin{icmlauthorlist}
\icmlauthor{Luke Metz}{goo}
\icmlauthor{Niru Maheswaranathan}{goo}
\icmlauthor{Jonathon Shlens}{goo}
\icmlauthor{Jascha Sohl-Dickstein}{goo}
\icmlauthor{Ekin D. Cubuk}{goo}

\end{icmlauthorlist}

\icmlaffiliation{goo}{Google Brain}

% \icmlcorrespondingauthor{Luke Metz}{lmetz@google.com}
\icmlcorrespondingauthor{}{lmetz@google.com}

\icmlkeywords{empty}

\vskip 0.3in
]

\printAffiliationsAndNotice{}  % leave blank if no need to mention equal contribution

\begin{abstract}
State-of-the art vision models can achieve superhuman performance on image classification tasks when testing and training data come from the same distribution. However, when models are tested on corrupted images (e.g. due to scale changes, translations, or shifts in brightness or contrast), performance degrades significantly.
Here, we explore the possibility of meta-training a learned optimizer that can train image classification models 
such that they are 
robust to common image corruptions.
Specifically, we are interested training models that are more robust to noise distributions not present in the training data.
We find that a learned optimizer meta-trained to produce models which are robust to Gaussian noise trains models that are more robust to Gaussian noise at other scales compared to traditional optimizers like Adam.
The effect of meta-training is more complicated when targeting a more general set of noise distributions, but led to improved performance on half of held-out corruption tasks.
Our results suggest that meta-learning provides a novel approach for studying and improving the robustness of deep learning models.
\end{abstract}

\section{Introduction}
Modern deep learning algorithms are exceptional at interpolation.
For example, they can achieve superhuman performance on image classification tasks when tested on the same distribution of images that they were trained on~\citep{karpathy2011lessons, krizhevsky2012imagenet, huang2018gpipe}.
When these models are evaluated on images that are even slightly perturbed, however, their performance often degrades catastrophically~\citep{dodge2017study,hendrycks2018benchmarking,azulay2018deep,rosenfeld2018elephant}.

A common way of increasing the robustness of deep learning algorithms is to apply perturbations to images during training~\citep{simard2003best,cubuk2018autoaugment}.
Although models trained with certain image perturbations become more robust to the specific perturbations they were trained with, they remain vulnerable to most other kinds of noise distributions~\citep{dodge2017study,hendrycks2018benchmarking,azulay2018deep,geirhos2018imagenet}.

In this work, we explore the effects of the optimization algorithm on robustness.
Specifically, we employ meta-learning to learn an optimizer designed specifically to 
produce models which 
perform well on corrupted images.
The meta-learning framework consists of two nested learning problems.
In the inner-problem, a learned, parametric optimizer trains a model, making use of gradients computed only on \textit{clean} training data.
The outer-problem involves training the parameters of the optimizer so that the model trained in the inner-loop has a low outer-loss.
In this work, we employ outer-losses based on validation performance on corrupted images.
We find that the learned optimizers produce models which are not only robust to the noise distribution used in outer-training, but, in some cases, are also more robust to additional noise distributions as well.

\section{Methods}
In this work we train task specific learned optimizers. Consider an inner-problem classifier with logits $f(x; w)$, parameters $w$, and input minibatch $x$. In this work, $f$ is a 4 layer CNN and $x$ is a minibatch of data sampled from Cifar10~\citep{krizhevsky2009learning}. Training this model can be expressed recursively:
\begin{align}
w^{\pp{t+1}} = w^{\pp{t}} - \text{U}\left(
 \nabla_{w^{\pp{t}}}l(y_{\text{train}}, f(x_{\text{train}};w^{\pp{t}})), \dots
\right),
\end{align}
where $y$ are the prediction targets, and $l$ is the cross entropy loss. The ellipses ($\dots$)
denote potential additional features passed to the update rule (e.g. momentum values).
An example of an update function $\text{U}$ is SGD, which can be expressed as: $U_{sgd}(g; \alpha) = \alpha g$ where $g$ is the gradient of the inner-loss, and the learning rate $\alpha$ is the single outer-parameter.
In this work, we introduce more complex update functions $U_{\text{meta}}(g, ...; \theta)$ with many outer-parameters $\theta$.

To evaluate inner-problem performance, we often use held out validation data. In this work, we additionally want to be robust to different kinds of corruptions. As such, to compute the outer-objective at inner-iteration $t$ we compute: $l(y_{\text{valid}}, f(\text{n}(x_{\text{valid}}); w^{\pp{t}}))$, where $\text{n}(\cdot)$ is a function which injects noise.
We emphasize that during outer-training of the optimizers, $\text{n}(\cdot)$ is not used to train the inner-model, only to evaluate it through the outer-objective. 
In some experiments, we do apply the learned optimizer to noised data after it has been outer-trained.

To find the outer-parameters, $\theta$, we optimize for performance of the meta-objective (noised validation loss) with a corruption chosen from the meta-training corruption set. In
all experiments
we employ truncated evolutionary strategies for training. While it is possible to use gradients, the estimators can be very high variance \citep{metz2018learned}.

In this work, we parameterize our learned optimizer similar to \citep{metz2018learned}, employing a small fully connected network that operates on each inner-parameter independently (with the exception of some cross parameter normalization, described in Appendix \ref{sec arch}). 
This parameterization leverages existing features from optimization (such as momentum at different scales \citep{lucas2018aggregated})
and is flexible enough to express common 
regularization techniques, such as weight decay or learning rate decay, since weight value and timestep are included as input features. See Appendix \ref{app:opt_details} for more information on the update rule parameterization and outer-training.

\section{Related Work}
Recent work highlights the contrast between the human visual system and artificial neural networks (ANN), by looking at commonplace corruptions of images. \citet{geirhos2018imagenet} reports that CNN rely much more on texture than shape, relative to humans. They find that data augmentation via style transfer can help ANNs focus more on shape, which leads to improved robustness to the Common Corruptions Benchmark~\citep{hendrycks2018benchmarking}. \citet{dodge2017study} report that while ANNs and humans perform comparably well on clean, high quality images; ANNs perform significanly worse on distorted images. They also report that errors made by humans and ANNs show little correlation (though other work has found surprising similarities in errors \citep{elsayed2018adversarial}). \citet{azulay2018deep} show that ANNs are not robust to geometric transformations of objects either, such as translations and scale changes. 

On the other hand, \citet{gilmer2018adversarial,fawzi2018adversarial,ford2019adversarial} show that robustness to commonplace corruptions and worst-case corruptions (such as adversarial examples~\citep{Szegedy14}) are directly related. \citet{cubuk2017intriguing} find that the sensitivity of ANNs to distortions at the input has a universal functional form across machine learning models, caused by a lack of correlation between outputs for different classes.  

Meta-learning is a general term often used to describe learning some aspects of a learning algorithm. Early work in this area is from \citet{schmidhuber1987evolutionary} which involves self-referential algorithms. Optimizer learning has been first been studied in \citep{bengio1990learning, bengio1992optimization} and then advanced with more complex parametric update rules and inner-models \citep{andrychowicz2016learning,chen2016learning,li2016learning,wichrowska2017learned, Bello17, metz2018learned}. In this work, we target an objective (validation loss on a noised image distribution) different than that used at training time (training loss). This idea has been explored in the context of validation loss, \citep{metz2018learned} as well as in unsupervised learning \citep{metz2018meta} and in reinforcement learning \citep{houthooft2018evolved}.

\begin{table*}[ht]
\vskip -0in
\begin{center}
\centerline{\includegraphics[width=0.9\textwidth]{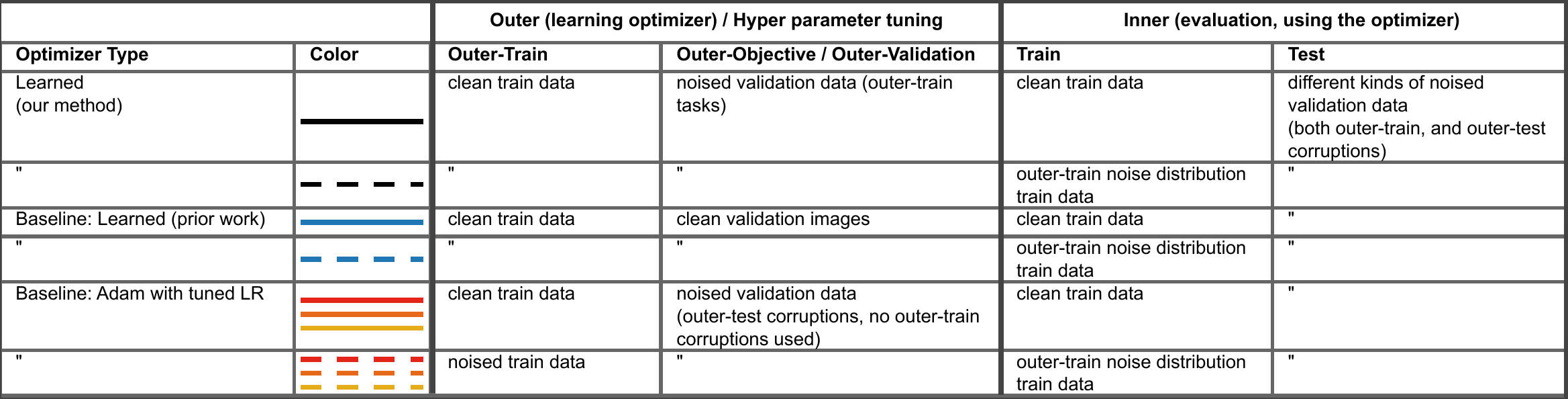}}
\caption{
List of different optimizers tested. Color denotes optimizer type.
A learned optimizer outer-trained on noisy validation images, our method (black), a baseline learned optimizer following \citep{metz2018learned} (blue), and LR tuned Adam (red through orange).
Solid lines denote evaluations by training a model on clean data, and then evaluating on a different noise distribution. Dashed lines denote evaluations when training on the outer-train noise distribution and evaluated on either the same, or different noise distribution.\label{tab optimizer list}}
\end{center}
\vskip -0.3in
\end{table*}

\begin{figure*}[ht]
\vskip -0.0in
\begin{center}
\centerline{\includegraphics[width=\textwidth]{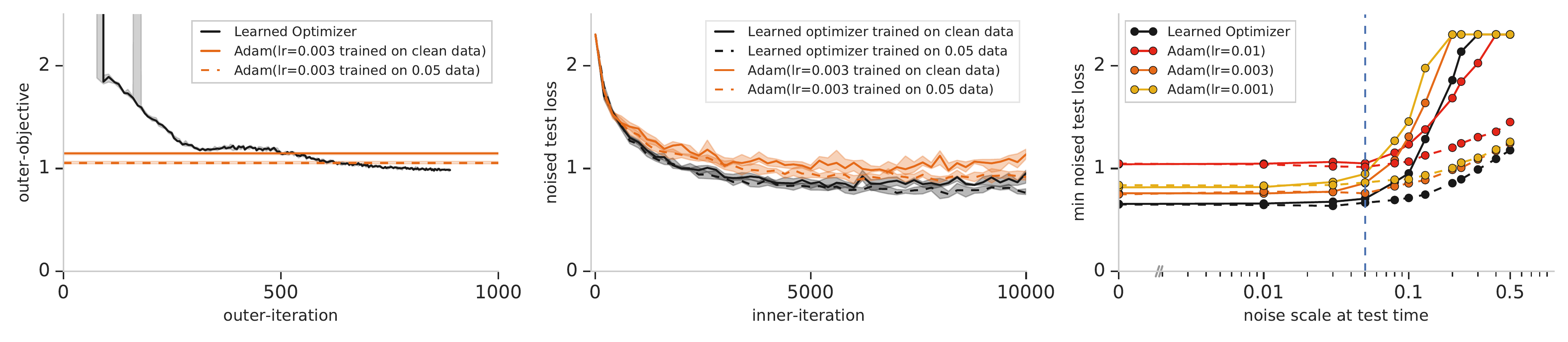}}
\caption{\textbf{(a)} Outer-training curves.
Each point represents an average of 10 inner-training runs using the learned optimizer for 10k iterations. We plot average outer-loss over the course of inner-training. Horizontal lines denote the best of learning rate tuned Adam, inner-trained on clean data (solid), and Adam inner-trained on noised data (dashed).
\textbf{(b)} Test performance on 0.05 noised images (same as outer-train noise amount).
Our learned optimizer on clean data outperforms both Adam on clean data (solid yellow), and Adam trained on noised data (dashed yellow).
\textbf{(c)} Test loss evaluated on varying amounts of noise. The model trained with the learned optimizer achieves a lower loss than all models trained with Adam at all noise scales. Solid lines denote clean data used for inner-training while dashed lines indicate training on 0.05 noised data. The vertical dashed line denotes the outer-training noise value.
\label{fig:noisedfig}
}
\end{center}
\vskip -0.3in
\end{figure*}

\section{Experiments}
We perform experiments on two types of noise distributions. First, we explore a corruption distribution consisting of different amounts of Gaussian noise added to the input image. Second, we explore a noise distribution based on the \citep{hendrycks2018benchmarking} corruption benchmark. We select an outer-train set of corruptions and test our method on held out corruptions. In all cases the inner-model, the model being trained by an optimizer, consists of a 4 layer CNN on Cifar10. All values reported are cross entropy loss calculated on test images.

To aid in clarity, we color code our experimental setup in Table \ref{tab optimizer list}.
For both experiments we train a learned optimizer. Our contribution, shown in black, is a learned optimizer outer-trained to perform well on noised validation data.
At evaluation time, we can assess performance by inner-training on either clean data (to match how it was outer-trained), or on noised data, and testing performance of the trained model on different noise distributions.
For the corruption data, to help isolate the effects of a more powerful optimizer, and of outer-training to target model robustness, we employ a second learned optimizer (blue) where we outer-train targeting clean validation images.

For both experiments we include Adam~\citep{kingma2014adam} baselines, with learning rate tuned over $10^{0.5 n}, n\in\mathbb Z$,
outer-trained on both clean (solid), and noised (dashed) data matching the outer-training corruption distribution.
To match standard hyperparameter tuning, we select the learning rate base on the target noise distribution, as opposed to the outer-train noise distribution.

\subsection{Gaussian Noise}

In this section, we train a learned optimizer to perform well on validation images (scaled 0-1) that have 0.05 per pixel Gaussian noise added to them. In Figure \ref{fig:noisedfig}a, we show outer-training curves. We find that our learned optimizer starts to outperform the learning rate tuned Adam after 500 outer-iterations and Adam inner-trained on noisy data after 600 outer-iterations. In Figure \ref{fig:noisedfig}b, we show inner-training of our learned optimizer evaluated on the noise distribution used at outer-training time. We present 2 baselines: first the learning rate tuned Adam trained on clean data, as well as the learning rate tuned Adam on the 0.05 noised-training data.
We find that despite never seeing noised data at inner-training time, our learned optimizer can outperform Adam specifically trained at this noise level.

In Figure \ref{fig:noisedfig}c we show outer-generalization outside the outer-training distribution. We present 2 settings of inner-training: training on clean data (solid) and 0.05 noised data (dashed). On clean data, our learned optimizer outperforms the clean Adam baselines but does not outperform Adam on noised data after 0.08 noise. When training on noised data, we find considerable improvements in robustness and outperform all other models. 
This is particularly surprising as this learned optimizer has never seen noised inner-training data at outer-training time.
Ideally we would like the leaned optimizer to outperform Adam when inner-trained on noisy data. While this is true for 0.05 noise, (solid black is lower than dashed yellow), this does not hold at higher noise levels.

\subsection{Novel corruption types}

\begin{figure*}[ht]
\vskip 0.0in
\begin{center}
\centerline{\includegraphics[width=\textwidth]{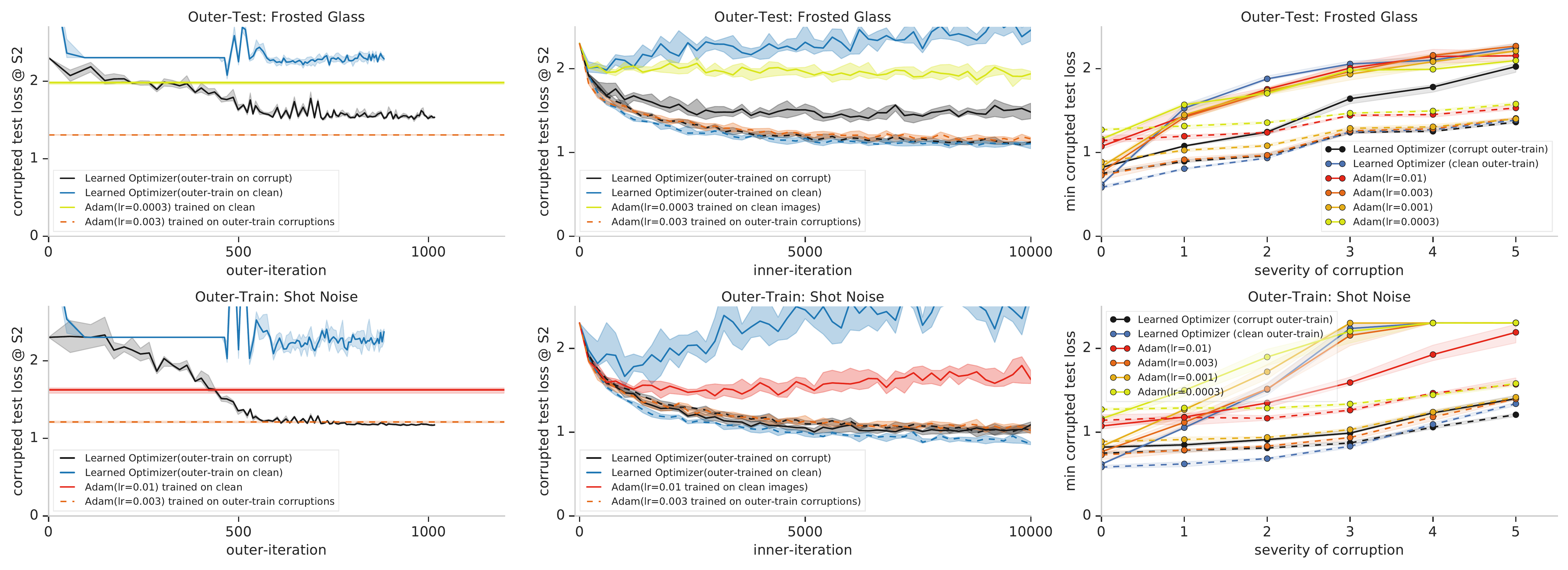}}
\caption{In each row we present a different corruption type. In \textbf{top} we show the outer-test corruption frosted glass blur, and in the \textbf{bottom} we show an outer-train corruption shot noise.
In \textbf{Column a}, we show evaluations of a given optimizer against these corruptions over the course of outer-training.
In \textbf{Column b} we perform inner-training of the previous two optimizers using both clean data (solid) and an inner-training set consisting of the same corruptions in the outer-training distribution (dashed). We find our learned optimizer outer-trained to be robust to corruption outperforms LR tuned Adam on clean data in both cases, but does not outperform inner-training on corrupted images (dashed lines). 
In \textbf{Column c} we show test performance after inner-training on clean data (solid) or noised data matching the meta-training corruption set (dashed). We then evaluate each model with different severities of the target corruption. We find our learned optimizer generalizes outside the severity region we outer-train on (severities 1-3) and outperforms Adam (red-orange) and the learned optimizer baseline (blue) on clean data.
Once again we find the inner-training on noise data (dashed) dramatically increases performance with both Adam and our learned optimizers. In all cases, we only inner-train on the outer-train corruption distribution and not on the corruption being evaluated. 
\label{fig:corruptfig}
}
\end{center}
\vskip -0.3in
\end{figure*}

In this section we explore the effects of transferring between \textit{different kinds} of corruptions.
We take the set of corruptions proposed in~\citet{hendrycks2018benchmarking}, divide the set of the nine corruptions (excluding JPEG corruption) into an outer-train set consisting of 7 training corruptions  (Gaussian noise, shot noise, impule noise, defocus blur, zoom blur, brightness, and contrast), and an outer-test set consisting of 2 corruptions (frosted glass blur and fog).
We outer-train only monitoring 2 train corruptions and the 2 test corruptions for computational reasons. In Figure \ref{fig:corruptfig}, we show two of the better performing corruptions (frosted glass, and shot noise) and provide the other two (fog, and brightness) in Appendix \ref{additional_corruption}.
As an additional baseline, to isolate the effect of having a better optimizer as opposed to an outer-training against a corruption objective, we also outer-train an optimizer targeting performance on clean validation images.

We find the performance of our learned optimizer varies dramatically across both the outer-train, and outer-test corruptions. We find our learned optimizer outer-trained for robustness, when inner-trained on clean data outperformed both the baseline learned optimizer, and the Adam when also inner-trained on clean data in all cases except the brightness corruption.
Once again we find inner-training on the outer-train corruption distribution helps dramatically for both Adam, and both learned optimizers. In the Appendix, we find for fog, and brightness, our baseline learned optimizer outperforms both our learned optimizer, and Adam.

\section{Discussion}

In this work we demonstrate the use of meta-learning to outer-train 
optimizers that produce robust classifiers.
While small scale, we see our results as the first step towards achieving this goal in real world settings. In this work, we present two extremes of how to parameterize optimizers: our MLP learned optimizer, and the learning rate tuned Adam. The Adam parameterization used in this work is limited, as it isn't able to make use of learning rate decay and regularizers like our learned optimizer. Designing better inductive biases and parameterizations for robustness on either end of the spectrum would be greatly beneficial. For example, the use of other regularizers (e.g. dropout), or data augmentation techniques would likely improve both our baseline and the learned optimizers.

In this work, we make the simplifying assumption for our learned optimizers that we are always inner-training on clean data. This choice defines a specific experimental paradigm.
We outperform the hand designed optimizers in most cases when the hand designed optimizers abide by this paradigm (Adam trained on clean data, solid lines).
When we break this experimental setup and train on noised data (dashed) we achieve much better performance with both Adam and our learned optimizers. 
Future work involves further exploring the impact of the training distribution on the meta-learning procedure.
We could, for example, inner-train on a distribution of corruptions, train an optimizer to target a different set, and outer-test on a third set.

A limitation of meta-learning is the need for a distribution of corruptions. We have found the existing set of 9 corruptions presented in~\citet{hendrycks2018benchmarking} are quite different in nature. This makes outer-generalization to unseen corruptions challenging.
Techniques such as meta-unsupervised learning \citep{hsu2018unsupervised} could be used to build heuristic corruption types to train on with the hope that the learned optimizer would transfer.

\section*{Acknowledgements}

We would like to thank Justin Gilmer for discussion on this project as well as the rest of the Brain Team.

\bibliography{main}

\begin{thebibliography}{31}
\providecommand{\natexlab}[1]{#1}
\providecommand{\url}[1]{\texttt{#1}}
\expandafter\ifx\csname urlstyle\endcsname\relax
  \providecommand{\doi}[1]{doi: #1}\else
  \providecommand{\doi}{doi: \begingroup \urlstyle{rm}\Url}\fi

\bibitem[Andrychowicz et~al.(2016)Andrychowicz, Denil, Gomez, Hoffman, Pfau,
  Schaul, and de~Freitas]{andrychowicz2016learning}
Andrychowicz, M., Denil, M., Gomez, S., Hoffman, M.~W., Pfau, D., Schaul, T.,
  and de~Freitas, N.
\newblock Learning to learn by gradient descent by gradient descent.
\newblock In \emph{Advances in Neural Information Processing Systems}, pp.\
  3981--3989, 2016.

\bibitem[Azulay \& Weiss(2018)Azulay and Weiss]{azulay2018deep}
Azulay, A. and Weiss, Y.
\newblock Why do deep convolutional networks generalize so poorly to small
  image transformations?
\newblock \emph{arXiv preprint arXiv:1805.12177}, 2018.

\bibitem[Bello et~al.(2017)Bello, Zoph, Vasudevan, and Le]{Bello17}
Bello, I., Zoph, B., Vasudevan, V., and Le, Q.
\newblock Neural optimizer search with reinforcement learning.
\newblock 2017.
\newblock URL \url{https://arxiv.org/pdf/1709.07417.pdf}.

\bibitem[Bengio et~al.(1992)Bengio, Bengio, Cloutier, and
  Gecsei]{bengio1992optimization}
Bengio, S., Bengio, Y., Cloutier, J., and Gecsei, J.
\newblock On the optimization of a synaptic learning rule.
\newblock In \emph{Preprints Conf. Optimality in Artificial and Biological
  Neural Networks}, pp.\  6--8. Univ. of Texas, 1992.

\bibitem[Bengio et~al.(1990)Bengio, Bengio, and Cloutier]{bengio1990learning}
Bengio, Y., Bengio, S., and Cloutier, J.
\newblock \emph{Learning a synaptic learning rule}.
\newblock Universit{\'e} de Montr{\'e}al, D{\'e}partement d'informatique et de
  recherche op{\'e}rationnelle, 1990.

\bibitem[Chen et~al.(2016)Chen, Hoffman, Colmenarejo, Denil, Lillicrap,
  Botvinick, and de~Freitas]{chen2016learning}
Chen, Y., Hoffman, M.~W., Colmenarejo, S.~G., Denil, M., Lillicrap, T.~P.,
  Botvinick, M., and de~Freitas, N.
\newblock Learning to learn without gradient descent by gradient descent.
\newblock \emph{arXiv preprint arXiv:1611.03824}, 2016.

\bibitem[Cubuk et~al.(2017)Cubuk, Zoph, Schoenholz, and
  Le]{cubuk2017intriguing}
Cubuk, E.~D., Zoph, B., Schoenholz, S.~S., and Le, Q.~V.
\newblock Intriguing properties of adversarial examples.
\newblock \emph{arXiv preprint arXiv:1711.02846}, 2017.

\bibitem[Cubuk et~al.(2018)Cubuk, Zoph, Mane, Vasudevan, and
  Le]{cubuk2018autoaugment}
Cubuk, E.~D., Zoph, B., Mane, D., Vasudevan, V., and Le, Q.~V.
\newblock Autoaugment: Learning augmentation policies from data.
\newblock \emph{arXiv preprint arXiv:1805.09501}, 2018.

\bibitem[Dodge \& Karam(2017)Dodge and Karam]{dodge2017study}
Dodge, S. and Karam, L.
\newblock A study and comparison of human and deep learning recognition
  performance under visual distortions.
\newblock In \emph{Computer Communication and Networks (ICCCN), 2017 26th
  International Conference on}, pp.\  1--7. IEEE, 2017.

\bibitem[Elsayed et~al.(2018)Elsayed, Shankar, Cheung, Papernot, Kurakin,
  Goodfellow, and Sohl-Dickstein]{elsayed2018adversarial}
Elsayed, G., Shankar, S., Cheung, B., Papernot, N., Kurakin, A., Goodfellow,
  I., and Sohl-Dickstein, J.
\newblock Adversarial examples that fool both computer vision and time-limited
  humans.
\newblock In \emph{Advances in Neural Information Processing Systems}, pp.\
  3910--3920, 2018.

\bibitem[Fawzi et~al.(2018)Fawzi, Fawzi, and Fawzi]{fawzi2018adversarial}
Fawzi, A., Fawzi, H., and Fawzi, O.
\newblock Adversarial vulnerability for any classifier.
\newblock \emph{arXiv preprint arXiv:1802.08686}, 2018.

\bibitem[Ford et~al.(2019)Ford, Gilmer, Carlini, and
  Cubuk]{ford2019adversarial}
Ford, N., Gilmer, J., Carlini, N., and Cubuk, D.
\newblock Adversarial examples are a natural consequence of test error in
  noise.
\newblock \emph{arXiv preprint arXiv:1901.10513}, 2019.

\bibitem[Geirhos et~al.(2018)Geirhos, Rubisch, Michaelis, Bethge, Wichmann, and
  Brendel]{geirhos2018imagenet}
Geirhos, R., Rubisch, P., Michaelis, C., Bethge, M., Wichmann, F.~A., and
  Brendel, W.
\newblock Imagenet-trained cnns are biased towards texture; increasing shape
  bias improves accuracy and robustness.
\newblock \emph{arXiv preprint arXiv:1811.12231}, 2018.

\bibitem[Gilmer et~al.(2018)Gilmer, Metz, Faghri, Schoenholz, Raghu,
  Wattenberg, and Goodfellow]{gilmer2018adversarial}
Gilmer, J., Metz, L., Faghri, F., Schoenholz, S.~S., Raghu, M., Wattenberg, M.,
  and Goodfellow, I.
\newblock Adversarial spheres.
\newblock \emph{arXiv preprint arXiv:1801.02774}, 2018.

\bibitem[Hendrycks \& Dietterich(2019)Hendrycks and
  Dietterich]{hendrycks2018benchmarking}
Hendrycks, D. and Dietterich, T.~G.
\newblock Benchmarking neural network robustness to common corruptions and
  surface variations.
\newblock \emph{International Conference on Learning Representations}, 2019.

\bibitem[Houthooft et~al.(2018)Houthooft, Chen, Isola, Stadie, Wolski, Ho, and
  Abbeel]{houthooft2018evolved}
Houthooft, R., Chen, R.~Y., Isola, P., Stadie, B.~C., Wolski, F., Ho, J., and
  Abbeel, P.
\newblock Evolved policy gradients.
\newblock \emph{arXiv preprint arXiv:1802.04821}, 2018.

\bibitem[Hsu et~al.(2018)Hsu, Levine, and Finn]{hsu2018unsupervised}
Hsu, K., Levine, S., and Finn, C.
\newblock Unsupervised learning via meta-learning.
\newblock \emph{arXiv preprint arXiv:1810.02334}, 2018.

\bibitem[Huang et~al.(2018)Huang, Cheng, Chen, Lee, Ngiam, Le, and
  Chen]{huang2018gpipe}
Huang, Y., Cheng, Y., Chen, D., Lee, H., Ngiam, J., Le, Q.~V., and Chen, Z.
\newblock Gpipe: Efficient training of giant neural networks using pipeline
  parallelism.
\newblock \emph{arXiv preprint arXiv:1811.06965}, 2018.

\bibitem[Karpathy(2011)]{karpathy2011lessons}
Karpathy, A.
\newblock Lessons learned from manually classifying cifar-10.
\newblock \emph{Published online at http://karpathy. github.
  io/2011/04/27/manually-classifying-cifar10}, 2011.

\bibitem[Kingma \& Ba(2014)Kingma and Ba]{kingma2014adam}
Kingma, D.~P. and Ba, J.
\newblock Adam: A method for stochastic optimization.
\newblock \emph{arXiv preprint arXiv:1412.6980}, 2014.

\bibitem[Krizhevsky \& Hinton(2009)Krizhevsky and
  Hinton]{krizhevsky2009learning}
Krizhevsky, A. and Hinton, G.
\newblock Learning multiple layers of features from tiny images.
\newblock Technical report, University of Toronto, 2009.

\bibitem[Krizhevsky et~al.(2012)Krizhevsky, Sutskever, and
  Hinton]{krizhevsky2012imagenet}
Krizhevsky, A., Sutskever, I., and Hinton, G.~E.
\newblock Imagenet classification with deep convolutional neural networks.
\newblock In \emph{Advances in Neural Information Processing Systems}, 2012.

\bibitem[Li \& Malik(2017)Li and Malik]{li2016learning}
Li, K. and Malik, J.
\newblock Learning to optimize.
\newblock \emph{International Conference on Learning Representations}, 2017.

\bibitem[Lucas et~al.(2018)Lucas, Sun, Zemel, and Grosse]{lucas2018aggregated}
Lucas, J., Sun, S., Zemel, R., and Grosse, R.
\newblock Aggregated momentum: Stability through passive damping.
\newblock \emph{arXiv preprint arXiv:1804.00325}, 2018.

\bibitem[Metz et~al.(2018)Metz, Maheswaranathan, Nixon, Freeman, and
  Sohl-Dickstein]{metz2018learned}
Metz, L., Maheswaranathan, N., Nixon, J., Freeman, C.~D., and Sohl-Dickstein,
  J.
\newblock Understanding and correcting pathologies in the training of learned
  optimizers.
\newblock \emph{arXiv preprint arXiv:1810.10180}, 2018.

\bibitem[Metz et~al.(2019)Metz, Maheswaranathan, Cheung, and
  Sohl-Dickstein]{metz2018meta}
Metz, L., Maheswaranathan, N., Cheung, B., and Sohl-Dickstein, J.
\newblock Meta-learning update rules for unsupervised representation learning.
\newblock \emph{ICLR}, 2019.

\bibitem[Rosenfeld et~al.(2018)Rosenfeld, Zemel, and
  Tsotsos]{rosenfeld2018elephant}
Rosenfeld, A., Zemel, R., and Tsotsos, J.~K.
\newblock The elephant in the room.
\newblock \emph{arXiv preprint arXiv:1808.03305}, 2018.

\bibitem[Schmidhuber(1987)]{schmidhuber1987evolutionary}
Schmidhuber, J.
\newblock \emph{Evolutionary principles in self-referential learning, or on
  learning how to learn: the meta-meta-... hook}.
\newblock PhD thesis, Technische Universit{\"a}t M{\"u}nchen, 1987.

\bibitem[Simard et~al.(2003)Simard, Steinkraus, Platt, et~al.]{simard2003best}
Simard, P.~Y., Steinkraus, D., Platt, J.~C., et~al.
\newblock Best practices for convolutional neural networks applied to visual
  document analysis.
\newblock In \emph{Proceedings of International Conference on Document Analysis
  and Recognition}, 2003.

\bibitem[Szegedy et~al.(2014)Szegedy, Zaremba, Sutskever, Bruna, Erhan,
  Goodfellow, and Fergus]{Szegedy14}
Szegedy, C., Zaremba, W., Sutskever, I., Bruna, J., Erhan, D., Goodfellow, I.,
  and Fergus, R.
\newblock Intriguing properties of neural networks.
\newblock In \emph{International Conference on Learning Representations}, 2014.
\newblock URL \url{http://arxiv.org/abs/1312.6199}.

\bibitem[Wichrowska et~al.(2017)Wichrowska, Maheswaranathan, Hoffman,
  Colmenarejo, Denil, de~Freitas, and Sohl-Dickstein]{wichrowska2017learned}
Wichrowska, O., Maheswaranathan, N., Hoffman, M.~W., Colmenarejo, S.~G., Denil,
  M., de~Freitas, N., and Sohl-Dickstein, J.
\newblock Learned optimizers that scale and generalize.
\newblock \emph{International Conference on Machine Learning}, 2017.

\end{thebibliography}
\bibliographystyle{icml2019}

\clearpage

\appendix
\section{Optimizer Details} \label{app:opt_details}
We briefly give an overview of our optimzier training details. The optimizer used in this work is similar to that used in \citep{metz2018learned}.

\subsection{Inner-Model}
The inner-model used in this work consists of a 4 layer convolutional neural network with ReLU activations. It contains hidden sizes of 32, 32, 64, 64 with strides 2,2,1,1. All layers use a kernel size of 3. The final layer is meaned spatially then passed into a linear projection to 10 units. We use cross entropy loss to train.

When outer-training our learned optimizer, we use clean Cifar10 data rescaled to fall between 0-1. Note that at evaluation time (after the model has been outer-trained) we also inner-train on noised data.

\subsection{Learned optimizer architecture}\label{sec arch}
The learned optimizer consists of a 1 hidden layer MLP that is shared across all units.
For each unit, we construct a feature vector containing a variety of features commonly used in hand designed optimizers \citep{wichrowska2017learned}. These include the gradient values, momentum values at 5 timescales, (0.5, 0.9, 0.99, 0.999, 0.9999), the current weights, the log absolute value of the weights. These values are then normalized by the second moment of each feature across each tensor. We include time based features consisting of $\text{sin}(\dfrac{st}{\pi})$ where $t$ is the current inner-training iteration, and $s$ is one of [2, 10, 20, 100, 200, 1000]. Additionally, a feature that is the log norm of each tensor value, and the log of the number of units in the tensor.

These features are all passed through a 1 hidden layer MLP with 32 units to produce 2 outputs: $a,b$. We combine them to produce a step as follows: $\Delta W = 0.001 a \text{exp}(0.001b)$. The form of this update can be thought of as learning a direction $a$, and a log step length $b$. We multiply by $0.001$ to ensure that the initial step size is stable and so that we do not initialize in an unstable outer-loss regime.

\subsection{Outer-training details}
We outer-train on a asynchronous, batched distributed cluster containing 256 workers and a batch size of 256. Each worker performs partial truncations and sends gradient information to a centralized learner. A worker then synchronizes weights, and proceeds from where the previous truncation left off. To account and mitigate truncation bias, we use an increasing schedule of truncation length that starts at 100 and linearly increases to 10k over 5k outer-iterations. Note that we never actually train until completion in any of our experiments. To prevent artifacts arising from the truncation schedule, we jitter this truncation amount by 20\% while training. If at any point the outer-loss is greater than 2 times the initial loss we stop the unroll, and reinitialize the inner-model randomly.

For a outer-gradient estimator, we make use of variational optimization. As shown in \citep{metz2018learned} we can use a reparametization based gradient (backaprop through unrolled training), or a gradient based on evolutionary strategies, or the combination of the 2. In this work, we only use the evolutionary strategies based estimator as it uses less ram with our naive implementation and is thus easier to work with given our computing infrastructure. We expect using the combined estimator would speed up outer-training. For lower variance evolutionary strategies gradients we make use of antithetical sampling with shared randomness where ever possible.

While progress has been made on increasing stability of learned optimizer training, not all random seeds converge. We use the outer-train loss to select the best model out of 4 random seeds for the corruptions experiments, and 3 random seeds for the Gaussian noise experiments.

\subsection{Outer-training task distribution: Gaussian experiments}
Our outer-objective for the Gaussian noise experiments consists of validation Cifar10 images corrupted with 0.05 Gaussian noise added to them.

\subsection{Outer-training task distribution: Corruption experiments}
Our outer-objective for the corruption experiments consists of sampling a severity amount, (1, 2, or 3), and a training corruption, (gaussian noise, shot noise, impulse noise, defocus blur, zoom blur, brightness and contrast). Each inner-training we sample a new augmentation to compute the meta-objective with. 

\section{Additional Corruption Plots} \label{additional_corruption}
\begin{figure*}[ht]
\vskip 0.0in
\begin{center}
\centerline{\includegraphics[width=\textwidth]{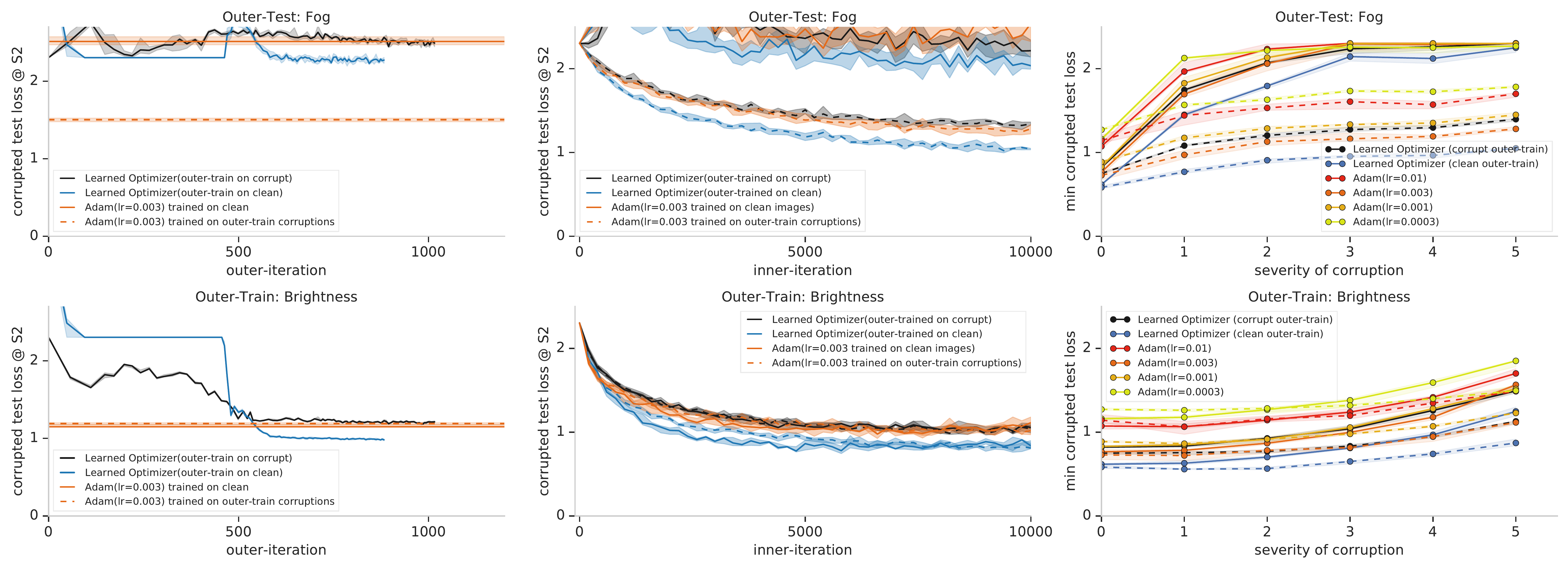}}
\caption{In each row we present a different corruption type. In \textbf{top} we show the outer-test corruption fog, and in the \textbf{bottom} we show an outer-train corruption brightness.
In \textbf{Column a}, we show evaluations of a given optimizer against these corruptions over the course of outer-training.
In \textbf{Column b} we perform inner-training of the previous two optimizers using both clean data (solid) and an inner-training set consisting of the same corruptions in the outer-training distribution (dashed).
In \textbf{Column c} we show test performance after inner-training on clean data (solid) or noised data matching the meta-training corruption set (dashed). We then evaluate each model with different severities of the target corruption.
}
\end{center}
\vskip -0.2in
\end{figure*}

\end{document}